\documentclass[journal]{IEEEtran}

\usepackage{cite}
\usepackage{hyperref}  
\usepackage{booktabs}  
\usepackage{graphicx}
\graphicspath{{figs/}}
\usepackage{tabularx}
\newcolumntype{L}{>{\raggedright\arraybackslash}X}
\usepackage{array} 
\usepackage{multirow}
\usepackage{makecell}
\usepackage{subfigure}
\usepackage{amsmath}

\usepackage[ruled,vlined]{algorithm2e}
\SetKwInOut{Parameter}{Parameters}

\usepackage{widetext}

\begin{document}

\title{Interpretable Reinforcement Learning Inspired by Piaget's Theory of Cognitive Development}

\author{Aref~Hakimzadeh,  
        Yanbo~Xue, 
        and~Peyman~Setoodeh
\thanks{A. Hakimzadeh and P. Setoodeh are with the School
of Electrical and Computer Engineering, Shiraz University, Shiraz,
Iran (e-mail: sa.hakimzadeh@shirazu.ac.ir; psetoodeh@shirazu.ac.ir).}
\thanks{Y. Xue is with the Career Science Lab, Beijing, China, and also with the Department of Control Engineering, Northeastern University 
Qinhuangdao, China (e-mail: yxue@careersciencelab.com).}
}

\maketitle

\begin{abstract}
Endeavors for designing robots with human-level cognitive abilities have led to different categories of learning machines. According to Skinner's theory, reinforcement learning (RL) plays a key role in human intuition and cognition. Majority of the state-of-the-art methods including deep RL algorithms are strongly influenced by the connectionist viewpoint. Such algorithms can significantly benefit from theories of mind and learning in other disciplines. This paper entertains the idea that theories such as language of thought hypothesis (LOTH), script theory, and Piaget's cognitive development theory provide complementary approaches, which will enrich the RL field. Following this line of thinking, a general computational building block is proposed for Piaget's schema theory that supports the notions of productivity, systematicity, and inferential coherence as described by Fodor in contrast with the connectionism theory. Abstraction in the proposed method is completely upon the system itself and is not externally constrained by any predefined architecture. The whole process matches the Neisser's perceptual cycle model. Performed experiments on three typical control problems followed by behavioral analysis confirm the interpretability of the proposed method and its competitiveness compared to the state-of-the-art algorithms. Hence, the proposed framework can be viewed as a step towards achieving human-like cognition in artificial intelligent systems.
\end{abstract}


\section{Introduction}
\label{intro}
The majority of the proposed RL algorithms in the literature have been built on the connectionism theory, which entertains the idea that learning is a result of forming associations between stimuli and response \cite{fodor1988connectionism}. Connectionism was first introduced by Thorndike in 1910, and as a result of this theory, neural network has become a prominent branch of artificial intelligence. Alternatively, the LOTH assumes that the linguistic structure is the basis of mind as well as the core of intuition, perception, and inference in human \cite{aydede2010language}. The LOTH was first introduced by Fodor in 1975 to address some of the unanswered issues with the connectionist viewpoint \cite{fodor1988connectionism} such as productivity, systematicity, and inferential coherence. Productivity refers to the ability of producing an infinite number of distinct representations in a system. Systematicity holds when the ability of a system to express certain propositions is intrinsically related to its ability to express some other propositions. In other words, systematicity is associated with interrelations among thoughts and inferences. Inferential coherence refers to a system's ability of producing original propositions. These issues motivated Fodor to go beyond the connectionist perspective and introduce the LOTH as the basis of thinking and learning.

As another well-known theory, schema theory was first developed by Piaget as a part of his cognitive development theory \cite{piaget2008psychology}, where schemas were introduced as the building blocks of human mind. He defined procedures that would modify schemas and lead to cognitive development. According to Piaget, ``\textit{schema is a cohesive repeatable action sequence possessing component actions that are tightly interconnected and governed by a core meaning}''. Assimilation and accommodation were proposed as two basic operations that human mind performs for inference and learning. Assimilation refers to using an existing schema to deal with a new object or situation, and accommodation refers to a procedure that considers the elements of a new situation that are either not contained in the existing schemas or contradict them \cite{anderson2000cognitive,hampson1996understanding}.

Behaviorism is another well-established theory about mind, which highlights the role of environmental factors in influencing behavior to the extent of almost ignoring innate or inherited factors \cite{skinner2011behaviorism}. This view emphasizes on the role of reward and punishment in the learning process, which is the main idea behind reinforcement learning. In this framework, structure of mind has been a less-explored topic and mind is usually treated as a black box. This viewpoint has led to a vast deployment of deep neural networks as black box universal function approximators in deep reinforcement learning algorithms.

\begin{figure}
\begin{center}
\includegraphics[width=0.8\linewidth]{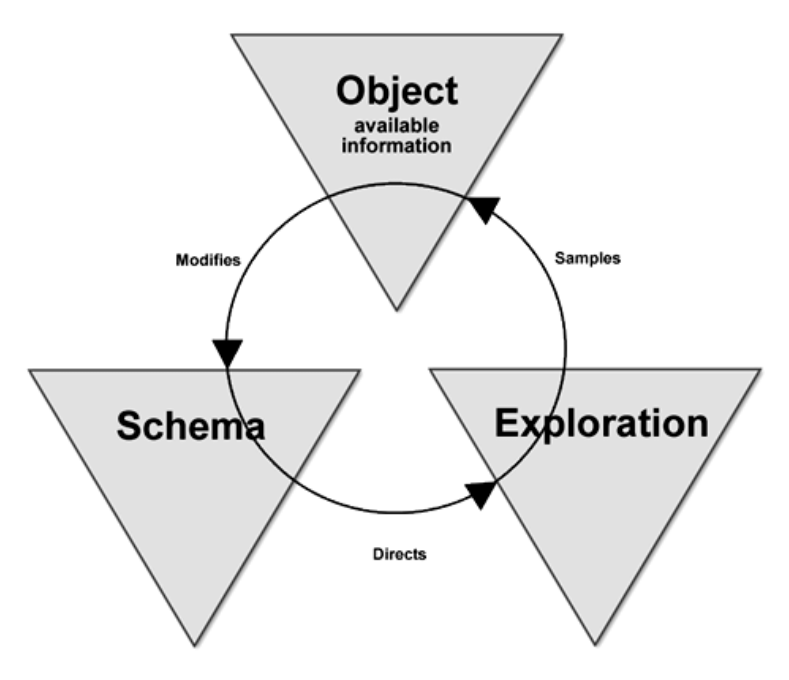}    
\caption{Neisser's perceptual cycle model: Directing, sampling, and internal modification shape a learning loop that encompasses both top-down and bottom-up learning.} 
\label{PCM}
\end{center}
\end{figure}

Another important topic that deserves attention in designing a general structure for mind is top-down versus bottom-up learning. Top-down learning as Gregory defines it \cite{gregory1970intelligent}, is acquiring explicit knowledge first, and then, learning implicit knowledge on that basis. Bottom-up learning developed by Gibson \cite{gibson1966senses}, is acquiring implicit knowledge first, and then, learning explicit knowledge on that basis \cite{sun2004top}. The debate on this issue and evidence provided for both views have led to the conclusion that the way human thinks benefits from both of these kinds of learning. Neisser proposed the perceptual cycle model (PCM) that regards learning as an infinite cycle that iterates between schema modification, directing the actions, and sampling from outside world (Fig. \ref{PCM}). According to Neisser's PCM, there is no need to implement mind's procedures as one-directional processes. This model includes both the bottom-up and top-down learning simultaneously \cite{hanson1996development,niesser1976cognition}. In addition to the PCM, Neisser has developed a schema theory in his book  in 1976 \cite{niesser1976cognition}, where he mentioned that ``\textit{generally a schema can be considered an organized mental pattern of thoughts or behaviors to help organize world knowledge}''. His interpretation of schema theory is helpful to design a cognitive system using reinforcement learning.

These mentioned theories provide valuable guidelines for designing the building blocks and the learning processes for cognitive systems \cite{cognitivecontrol2012, fatemi2017observability}. Schema theory provides a very promising complementary approach to behaviorism, which can address the inherent vagueness issue regarding the structure of mind. In addition, Neisser's PCM can help in designing a complete cycle for learning and improving the algorithm autonomy by decreasing the designer's intervention, for instance, in choosing the hyper-parameters. This paper proposes a framework for modelling and designing cognitive systems with emphasis on filling the gap between the state-of-the-art RL algorithms and these psychological and philosophical theories of cognition.

The rest of the paper is organized as follows. Next section briefly reviews the state-of-the-art RL algorithms and highlights the gap between the RL literature and the mentioned psychological and philosophical theories of mind and learning. In Section~3, the proposed framework for Abstraction via Options of Options (AO2) is explained as an attempt to fill this gap. Section~4 provides computer experiments on three typical case studies to evaluate the proposed method and provide a proof of concept. Section~5 is dedicated to behavioral analysis of the proposed algorithm. Finally, concluding remarks are provided in Section~6.

\section{Related Work}

Markov decision process (MDP) provides the mathematical framework for building most of the RL algorithms. An MDP is mathematically described by the tuple $(S, A, R, P, P_0,\gamma)$, where $S$ denotes the state space, $A$ denotes the action space, $R$ is a reward function such that $R : S \times A \rightarrow R$, $P$ denotes the state-action transition probability distribution function (PDF), $P : S \times S \times A \rightarrow [0,1]$, $P_0$ is the initial state PDF, $P_0 : S \rightarrow [0,1]$, and $\gamma \in [0,1]$ is a discount factor. Sutton defined an option as the tuple $(I_o, \pi_o, \beta_o)$, where $I_o$ is the option initiation set, $\pi_o : A \times S \rightarrow [0,1]$ is the option's policy, and $\beta_o : S \rightarrow [0,1]$ is its termination function \cite{sutton1999between}. The goal is to maximize the expected collected reward, which is called return. There are three major categories of RL algorithms: critic-only, actor-only, and actor-critic. Critic-only algorithms such as Q-learning explicitly learn an optimal value function. Actor-only algorithms such as policy gradient method explicitly learn an optimal policy using the advantage function. Actor-critic algorithms explicitly learn both an optimal value function and an optimal policy. Trust Region methods maximize the collected reward by controlling the policy change at each step using the Kulback-Leibler divergence (KLD).

By defining the options framework, RL researchers have tried to address the criticism regarding the incapability of such algorithms in mimicking the human learning process as well as success in tasks that require long-term reward controlling \cite{sutton1999between,precup2001temporal}. The theory of options has inspired many researches. Primal work on the theory of options was mainly focused on finding subgoals and modifying the policy upon them, and therefore, did not provide a complete learning cycle \cite{mcgovern2001automatic,menache2002q,silver2012compositional,
csimcsek2009skill,stolle2002learning,daniel2016probabilistic,konidaris2011autonomous,
kulkarni2016hierarchical,mankowitz2016adaptive,mann2015approximate,
niekum2013semantically,vezhnevets2016strategic}. Option-critic learning has emerged from combination of option finding and learning \cite{bacon2017option,harb2018waiting,klissarov2017learnings}. In option-critic learning, action was replaced by option in all computations, and options became the core of the learning process. Later, hierarchical architectures were adopted in the options framework \cite{botvinick2008hierarchical,nachum2018near}. Starting with a two-level architecture, the hierarchy was extended to K levels of abstraction \cite{riemer2018learning} by implementing K levels of options, where each layer provides the lower-level material for the next layer to act in accordance with. Inspired by cognitive development, the idea of intrinsic motivation was proposed to navigate the learning path \cite{dayan1993improving,stachenfeld2014design}. This idea can be naturally implemented using deep learning \cite{kulkarni2016hierarchical}. In \cite{schulman2017proximal}, Proximal Policy Optimization (PPO) algorithm was presented, which uses a surrogate objective function along with gradient-based algorithms. Using a surrogate objective function along with gradient-based optimization has led to the Proximal Policy Optimization (PPO) \cite{schulman2017proximal}. In \cite{zhang2019dac}, a variation of the option-critic algorithm was proposed that uses two distinct MDPs as master and slave to learn options.

Most of these algorithms were developed under the influence of the connectionism viewpoint. However, integration of the connectionism viewpoint with productivity, systematicity, and inferential coherence would be beneficial. Moreover, some algorithms may put more emphasis on one directional learning view, either top-down or bottom-up, while many real-world problems have shown that one of these theories alone cannot cover all aspects of the human learning.

\section{The Proposed AO2 Algorithm}

\begin{figure}
\begin{center}
\includegraphics[width=0.8\linewidth]{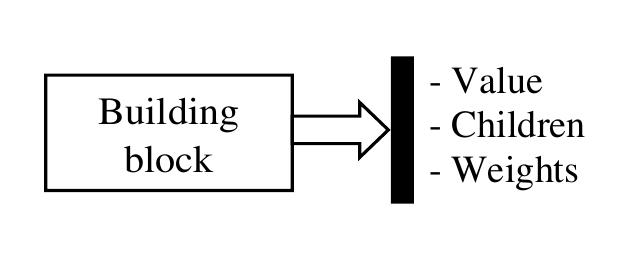}    
\caption{Building block of schemas (options).} 
\label{Option}
\end{center}
\end{figure}

\begin{figure}
\begin{center}
\includegraphics[width=0.8\linewidth]{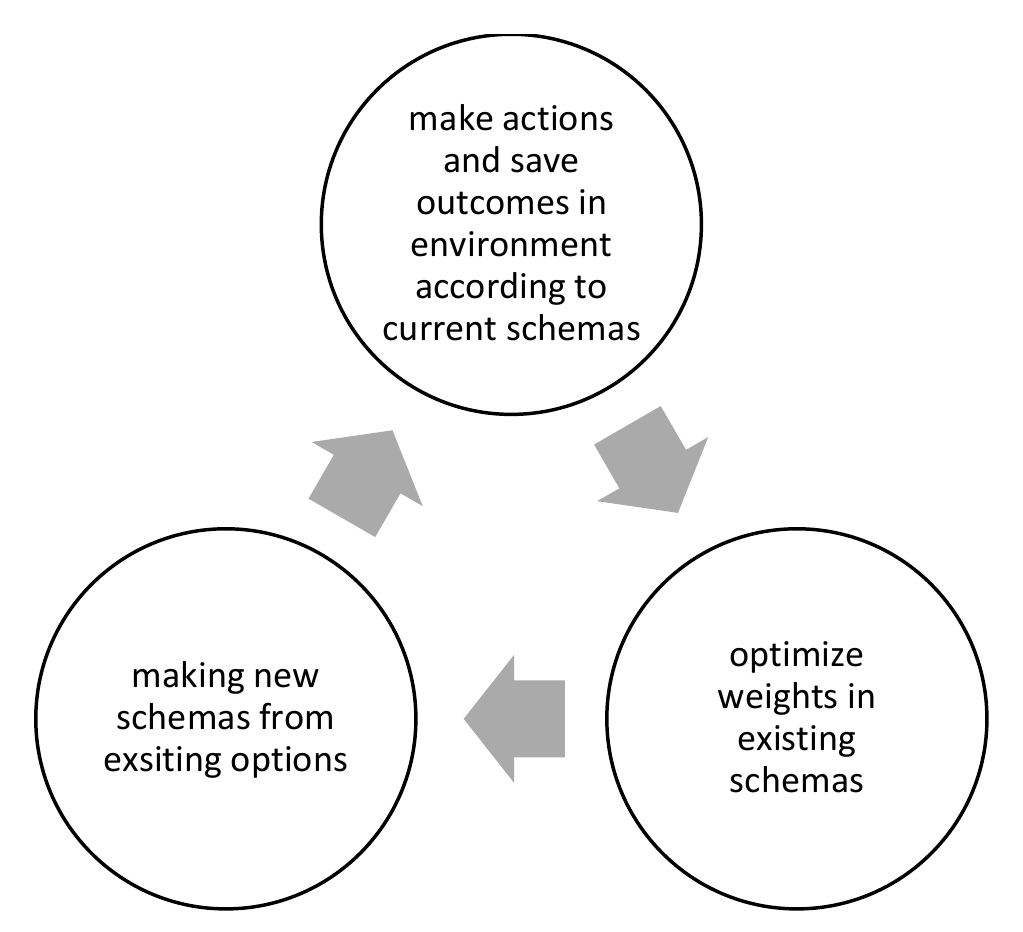}    
\caption{The learning process.} 
\label{Learn}
\end{center}
\end{figure}

Since the learning agent is meant to follow the LOTH, we need to propose a computational building block for schemas (i.e., options) as shown in Fig. \ref{Option}. The building block has an assigned value and some pointers to several other blocks of the same kind, called its children, and a weight for each child. Each option is the system's representation of a situation or event. The computational building block must have the following characteristics:
\begin{itemize}
\item \textbf{Schema as a mental unit}: A schema is defined as a tree of options, which are coherent in a special manner and lead to a specific goal. Each leaf of this tree contains a possible action for that system. Here, action refers to basic actions that each system can take by sending a stimulus to an output terminal (or an actuator) that lasts for one time step. The whole system contains a set of such schemas. Every option is able to possess every other option as its child. Moreover, each option can play the child's role for a number of other options.
\item \textbf{Similairty criterion}: 
Finding the most similar option to the current state is the basis of the proposed algorithm. A properly tailored similarity measure would provide a solution for both attention and feature selection problems. A simple candidate for such a measure can be defined as:
\begin{equation}
s = W^T |U - V_{O_{i}}|,    
\label{eq1}
\end{equation}
where $W$ denotes the system attention weights, $U$ is the sensory input, and $V_{O_{i}}$ is the value of the chosen option.
\item \textbf{Long and short term effects}: The trace length is a parameter that determines how many successive actions and their corresponding earned rewards must the system consider at each time-step for updating the activated schema's weights. The trace length is an essential design factor, which enables the system to deal with delayed rewards. The trace length allows the system to back track its functionality during a period of time instead of just considering one moment. In this way, credit assignment can be performed for long-term rewards by checking a sequence of selected actions not just the last one. 
\item \textbf{Inference}: 
Inference process in this system refers to finding the most similar schema to the current state of the system and tokenizing that schema: 
\begin{equation}
	\begin{split}
	& f = \underset{i}{argmin}~Similarity(currentState,Option_{i}.value) \\
	\end{split}.
\end{equation}
Substituting the similarity function (\ref{eq1}) in the above equation, we obtain:
\begin{equation}
	\begin{split}
& f = \underset{i}{argmin}~~ W^T |U - V_{O_{i}}| \\
	\end{split}.
\end{equation}
Activating the chosen schema leads to finding the best action, which earns the highest reward. Finding the best action in a schema occurs in a sequential manner through descendants of the chosen option. This process is mathematically formulated as the following bi-level optimization problem:
\begin{widetext}
\begin{equation}
	\begin{aligned}
	& \underset{j}{argmin} ~~~~~ w_{j}\\
	& \text{subject to:} ~~~  w_{j} \in \{Option_{s}.children\} \\
	& ~~~~~~~~~~~~~~~~~ s = \underset{k}{argmin}~similarity(currentState,option_{k}.value)  \\
	&  ~~~~~~~~~~~~~~~~~\text{subject to:} ~~~ option_{k} \in option_{f}.Descendants
	\end{aligned} 
\end{equation}
\end{widetext}
In this bi-level optimization, the lower level finds the most similar option, labeled as $option_{k}$, among the descendants of the tokenized option, which is labeled as $option_{f}$. Hence, the solution would be the most rewarded action of the most similar option. Considering the similarity criterion (\ref{eq1}) and denoting option by $O$, the above equation is rewritten as:
\begin{equation}
	\begin{aligned}
	& \underset{j}{argmin} ~~~~~ w_{j}\\
	& \text{subject to:} ~~~  w_{j} \in \{O_{s}.children\} \\
	& ~~~~~~~~~~~~~~~~~ s = \underset{k}{argmin}~~
  \|W^T (I - V_{O_{k}}) \|    \\
	&  ~~~~~~~~~~~~~~~~~\text{subject to:} ~~~ O_{k} \in O_{f}.Descendants
	\end{aligned} 
\end{equation}
\item \textbf{Learning process}: The learning process contains two procedures, which are iteratively executed (Algorithm \ref{alg:ao2}). The first one is a procedure that reforms the structure, where there is a contradiction or deficiency between the activated schema and the current state. In such cases, proper options are added to the tree structure or inappropriate options are removed from it (Algorithm \ref{alg:build}). Piaget defined this process as accommodation. The second procedure updates the tree weights to earn the maximum reward possible regarding the chosen action at the current state (Algorithm \ref{alg:reform}); Piaget named this process assimilation. Both of these procedures work based on the similarity between each option and the current state as well as the earned reward over a time interval determined by the trace length.
\item \textbf{Memory cleansing}: During the learning phase, each option is likely to find more similar options as its children than the existing ones. The algorithm's convergance rate can be increased by removing less alike children from an option. Another way of increasing the convergence rate is to keep only good actions for each option by removing actions with corresponding lower weights as the procedure goes on. Weights of children are determined after execution of each action considering the trace length and the collected rewards.
\end{itemize}
In summary, as demonstrated in Fig. \ref{Learn}, the whole process is based on iteratively sampling from world (i.e., environment), reforming structures, and modifying weights of schemas. Similarity is the basis of all operations in schemas, and the trace length is used to regulate the corresponding weights. 

\begin{algorithm}[h]
    \SetAlgoLined
    Initialization\;
    	
	\For{$i\leftarrow 1$ \KwTo $maxIterations$ }{
    \Repeat{done flag of the environment is 1}{
        selectedOption = FindMostSimilarOption(currentState)\;
		action = tokenizeMostEfficientAction(selectedOption)\;
		reward, nextState = environmrnt.step(action)\;
		reformStructures(Algorithm \ref{alg:build})\;
		applyRewards(Algorithm \ref{alg:reform})\;
    }
  }
 
\caption{\label{alg:ao2} Schema-based reinforcement learning}
\end{algorithm}

\begin{algorithm}[h]
    \SetAlgoLined
	selectedOption = FindMostSimilarOption(currentState)\;
    \For{$i\leftarrow 0$ \KwTo $Length(selectedOption.children)$ }{
    	newChildFlag = True\;
		\If{$similarity(selectedOption,currentState)> Threshold$}{
			newChildFlag = False\;
		}
    }
	\If{newChildFlag}{
		add a new child with value of \emph{currentState} to selectedOption\;
	}
 
\caption{\label{alg:build} Accomodation (Structure Reforming)}
\end{algorithm}

\begin{algorithm}[h]
    \SetAlgoLined
    \KwIn{pathHistory,~rewardsHistory}
    \Parameter{$\gamma$,~episodeLength}
    $c = 1$\;
    \For{$i\leftarrow 1$ \KwTo $pathHistory.Length~-episodeLength$ }{
    	$ c ~ *= \gamma $;\\
		\For{$j\leftarrow 1$ \KwTo $pathHistory[i].Length-1$ }{
			pathHistory[i].weights[j] += rewardsHistory[i] * c;
		}
    }
\caption{\label{alg:reform}Assimilation (Weight update)}
\end{algorithm}

\section{Experiments}

\begin{figure*}%[t!]
    \label{fig:envs}
    \centering
    \subfigure[Cart pole]{\label{fig:cart}\includegraphics[width=0.25\linewidth]{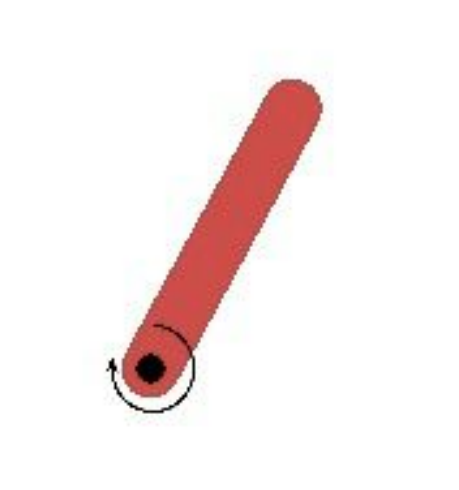}}    
    \subfigure[Pendulum]{\label{fig:pend}\includegraphics[width=0.30\linewidth]{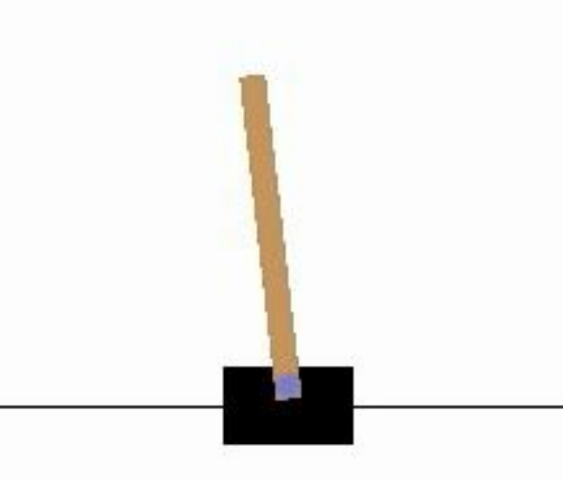}}    
    \subfigure[Acrobot]{\label{fig:acro}\includegraphics[width=0.30\linewidth]{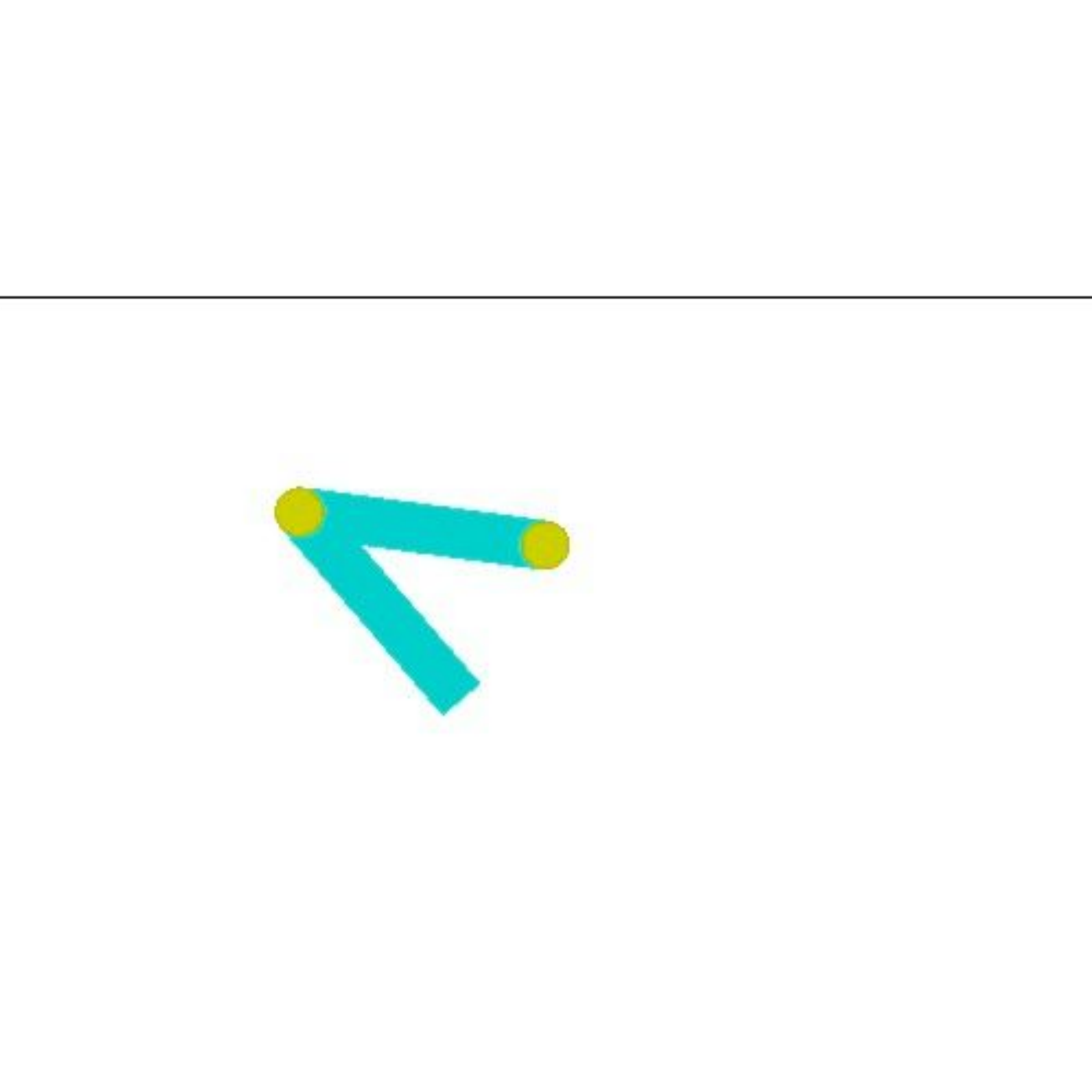}}    
    \caption{Gym environments.}
\end{figure*}

Three control tasks were chosen from the OpenAI Gym environments (Fig. \ref{fig:envs}) to evaluate the proposed algorithm. The proposed algorithm is compared with four well-known algorithms: PPO, Proximal Policy Option-Critic (PPOC), combination of Double Actor-Critic (DAC) and PPO, which is referred to as DAC+PPO, combination of DAC and Advantage Actor Critic (A2C), which is referred to as DAC+A2C. These algorithms were implemented according to the details provided in \cite{zhang2019dac}.

\begin{figure}%[t!]
    \label{fig:curves}
    \centering
    \subfigure[]{\label{fig:curve_1}\includegraphics[width=1\linewidth]{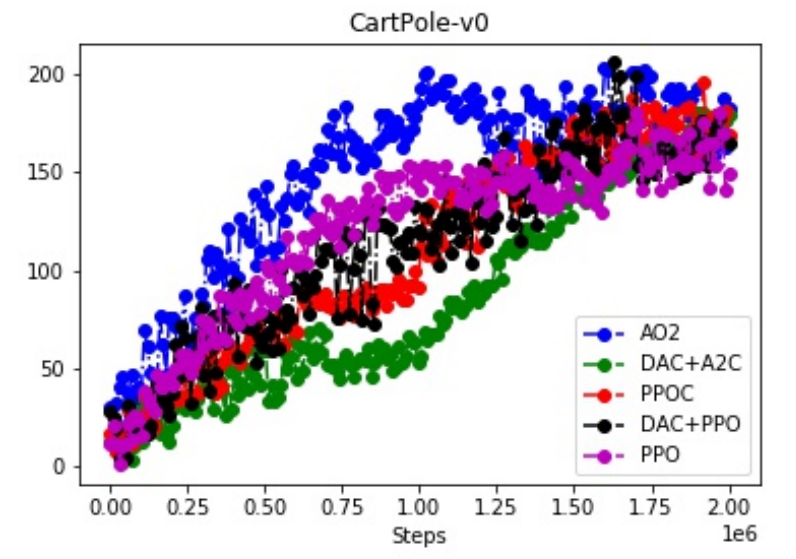}}    
    \subfigure[]{\label{fig:curve_2}\includegraphics[width=1\linewidth]{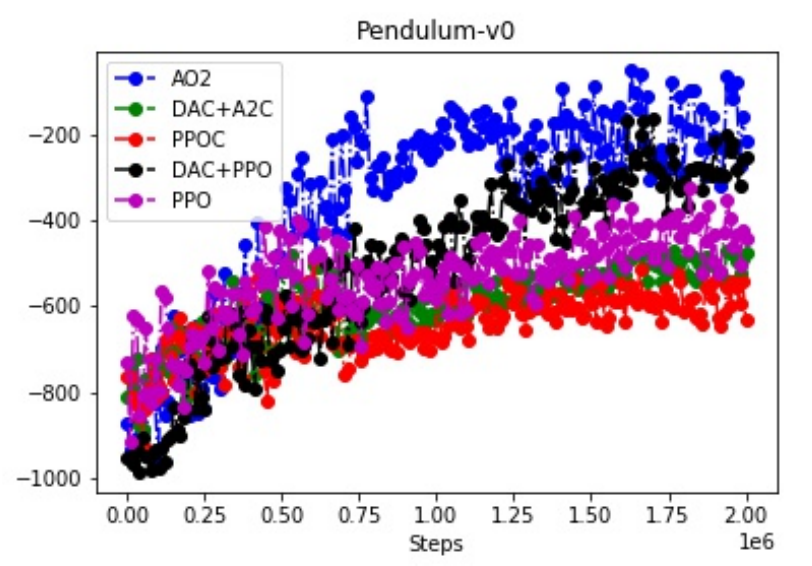}}    
    \subfigure[]{\label{fig:curve_3}\includegraphics[width=1\linewidth]{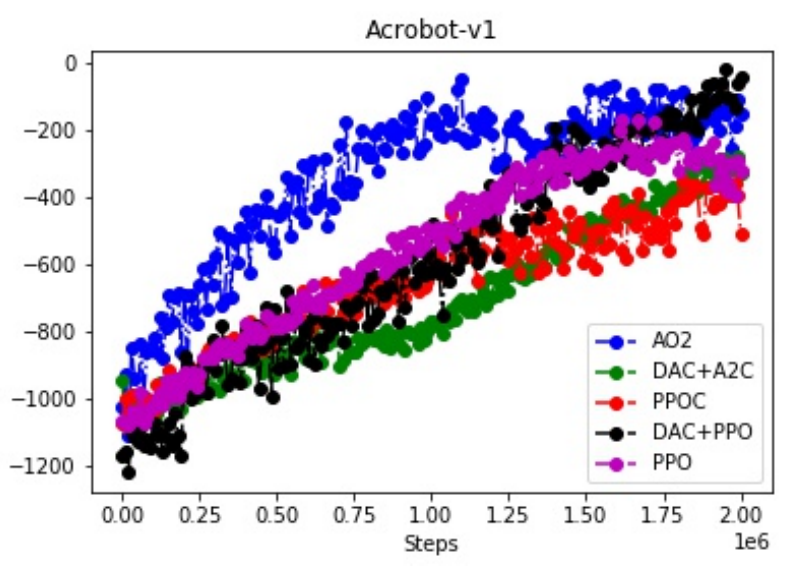}}    
    \caption{Learning curves of the implemented algorithms for different environments: (a) Cart Pole, (b) Pendulum, and (c) Acrobot.}
\end{figure}

\begin{figure}
\begin{center}
\includegraphics[width=1\linewidth]{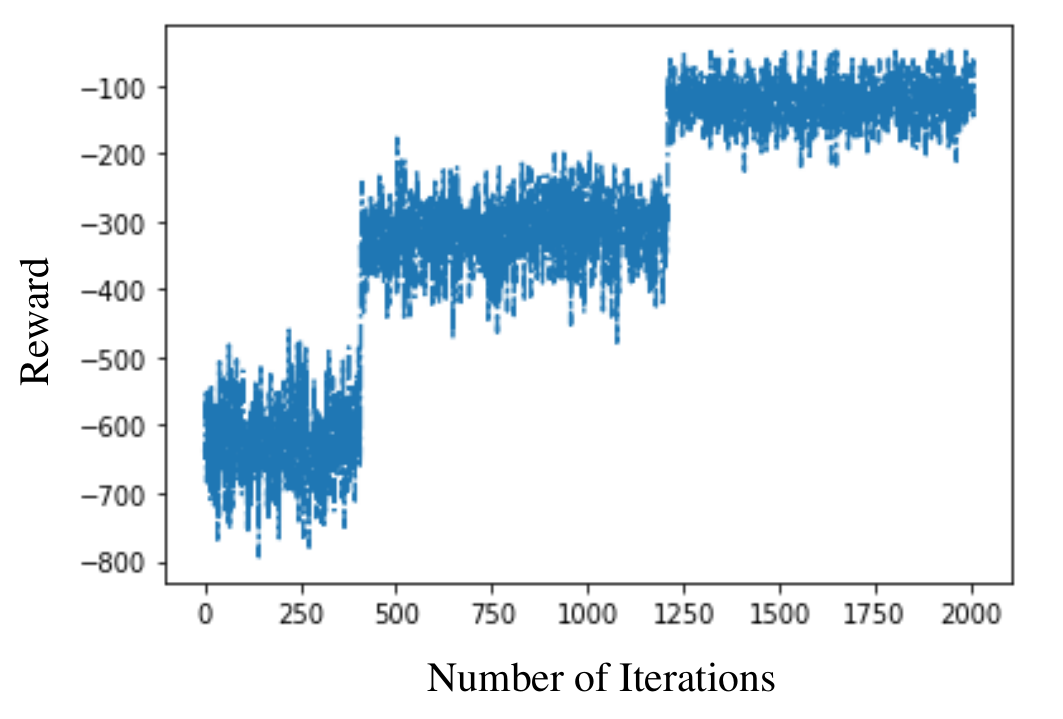}    
\caption{Reward mean value for every 200 steps during the training process for the pendulum problem.} 
\label{Result2}
\end{center}
\end{figure}

\begin{enumerate}
\item \textbf{Gym CartPole-v0}: The first set of experiments were performed on the CartPole-v0 (Fig. \ref{fig:cart}). Two actions are allowed in the cart pole environment; moving the cart to right or left. Reward is one for every taken step. Each episode terminates if the absolute value of the angle is more than 12$^\circ$, the cart position is not between $\pm$2.4, or 200 steps has been taken, then, the environment will be reset. The trace length was set to 5 and the threshold for similarity measure was 0.9 for both assimilation and accommodation. Both actions were assigned to each basic option. After training, the average reward was 197.2 and the best reward of 197.2 for 100 consecutive episodes. According to the OpenAI website, collecting this amount of reward can be considered as solving the problem. Size of the final option-set was 680 KB. The average and the best collective rewards are comapred for different algorithms in Table \ref{tab:table2} and the corresponding learning curves are shown in Fig. \ref{fig:curve_1}.
\item \textbf{Gym Pendulum-v0}: The second set of experiments were performed on the Pendulum-v0 environment (Fig. \ref{fig:pend}), which is more challenging than the cart pole. Control signal is a real number between $\pm$2, and reward takes continuous values such that when the pendulum stays at the top point, the agent receives zero reward and at other points it gets a negative reward proportional to the pendulum's distance from the top position. When the pendulum starts from the bottom, taking even the strongest possible action in one direction cannot bring it to the top point. In other words, the pendulum should swing to reach to the highest point. This process needs a longer trace length compared to the cart pole problem. The trace length was set to 18, the similarity criterion thresholds for assimilation and accommodation were both set to 0.99. After training, the system reached the average reward of -126.6 and the best reward of -196.5 for each episode for 100 consecutive episodes. Size of the final option set was about 4 MB. Fig. \ref{Result2} shows the average reward for every 200-step run during the training process. Two jumps are observed in Fig. \ref{Result2} that separate three phases. The number of distinct areas is equal to the number of times that a round of making new schemas, optimizing weights in existing schemas, and deciding according to the schemas occur in the simulation. In other words, jumps between each pair of these areas show effectiveness of Neisser's PCM. The average and the best collective rewards are comapred for different algorithms in Table \ref{tab:table2} and the corresponding learning curves are shown in Fig. \ref{fig:curve_2}.
\item \textbf{Gym Acrobot-v1}: The third set of experiments were performed on the Acrobot-v1 environment, which has two links hanging downward from a fixed joint (Fig. \ref{fig:acro}). The goal of this problem is that the end point of the lower link touches the horizontal line above the acrobot. The learning agent receives zero reward for every step except for the step of touching the horizontal line, where the reward is one. This is a difficult task to learn due to both reward sparsity and relatively large distance from the initial point to the desired point. The trace length was set to 80, the similarity criterion thereshold for assimilation and accommodation were both set to 0.99. After training, the system reached the average reward of -89.5 and the best reward of -77.12 for each episode for 100 consecutive episodes. The average and the best collective rewards are comapred for different algorithms in Table \ref{tab:table2} and the corresponding learning curves are shown in Fig. \ref{fig:curve_3}.
\end{enumerate}
Achieved results by AO2 in these three sets of experiments, which are compared with those of PPO, PPOC, DAC+A2C and DAC+PPO in Table \ref{tab:table2}, confirm the competitiveness of the proposed algorithm. Learning curves for all algorithms, which are depicted in Figure \ref{fig:curves} and show the gained reward during the learning process, demonstrate that AO2 has the fastest convergence among the implemented algorithms. 

\begin{table*}[h]
\begin{center}
    \caption{Comparison between AO2 and the state-of-the-art algorithms on three case-studies.} 
%{AO2 shows the best performance for the pendulum problem in the sense of achieving the best gained reward as well as the best average reward. For the cart pole problem, DAC+PPO achieves the best average reward, but AO2 gains the highest collected reward. For the acrobot problem, DAC+PPO achieves the best results and AO2 takes the second place.}
    \label{tab:table2}
  \begin{tabular}{c c c c c}
    \toprule
    & & CartPole-v0 & Pendulum-v0  & Acrobot-v1\\ 
    \midrule
    \multirow{2}{*}{AO2} & best & \textbf{197.2} & \textbf{-126.6} & -77.12\\
                         & average & 194.0 & \textbf{-196.6} & -89.5\\\hline
    \multirow{2}{*}{DAC+A2C} & best & 197.1 & -437.25 & -98.15\\
                         & average & 193.9 & -496.28 & -261.48 \\\hline
    \multirow{2}{*}{PPOC} & best & 197.1 & -513.84 & -104.58\\
                         & average & 193.7 & -623.17 & -268.17 \\\hline
    \multirow{2}{*}{DAC+PPO} & best & 197.0 & -141.18 & \textbf{-59.38}\\
                         & average & \textbf{196.1} & -204.81 &  \textbf{-69.24}\\\hline
    \multirow{2}{*}{PPO} & best & 196.8 & -164.19 & -101.63\\
                         & average & 191.5 & -241.16 & - 214.37\\
   \bottomrule
   \end{tabular}
\end{center}
\end{table*}

\section{behavioral Analysis}

In order to provide a thorough behavioral analysis of the proposed algorithm, additional experiments were performed considering the following characteristics:
\begin{itemize}
\item \textbf{Smooth actions}: Usually, human beings cannot shift their attention from one situation to another without being influenced by the former. Hence, each executive operation must take account of both the current inferred action and the previous executed actions. This limitation can be mathematically modelled as: 
\begin{eqnarray}
executable\ action &=& \alpha \times chosen\ action \nonumber \\  &+& (1-\alpha) \times last\ action, 
\end{eqnarray}
where $\alpha$ was set to 0.9 in this experiment. Implementing this idea on the pendulum environment resulted in the mean reward of -212 and the best reward of -167 in 10 experiments of 100 consecutive rounds. These results show that, at least for this task, the immediate computed action leads to a better performance.
\item \textbf{Exploration and exploitation}: The proposed algorithm is based on finding the best action for the most similar option to the current state. Hence, it eliminates the opportunity to try different things at a specific state. In other words, the proposed algorithm does not opt for exploration, and therefore, the system would be prone to getting stuck in local optima. To address this issue, the idea of the $\epsilon$-greedy policy is used and after finding the most similar option, the best action is chosen with a probability of $90\%$. In this way, the action will be randomly chosen from other available actions in that particular state with a probability of $10\%$. Increasing this probability improves the chance for exploration. Implementing this $\epsilon$-greedy policy for action selection resulted in the mean reward of -168 and the best reward of -135 in 10 experiments of 100 consecutive runs on the pendulum environment.
\item \textbf{Personal impression}: In the similarity measure formula, $W$ determines the relative importance of each sensory input in the process of finding the most similar state. Initializing $W$ as $\vec{1}$, the best weights for the sensory input array can be iteratively found based on the distance between the chosen option and the current state: 
\begin{equation}
W^{new} = W^{old} \pm \beta (U-V),    
\label{eq3}
\end{equation}
where $\beta=0.04$, $U$ is the sensory input, and $V$ is the chosen option. The $+$ sign is for the occasions that the moving average of gained rewards is increasing and the $-$ sign is for the opposite situations. This setting led to the mean reward of -241 and the best reward of -164 in 10 experiments of 100 consecutive runs on the pendulum environment. The point of this experiment was to demonstrate that different people may come to different conclusions in a specific situation and still perform equally well. Similarly, the implemented algorithm has acceptable performance with different weights. For instance, $W_{1} = [2.2\ 2.4\ 0.7]$ and $W_2 = [1.2\ 1.7\ 1.8]$ both led to the mean reward of -168 for the Pendulum problem.
\item \textbf{Hierarchical architecture}: Human cognition significantly benefits from hierarchies. As mentioned previously, each schema is modelled by a tree structure, where different leaves may have different depths with their corresponding different weights that can be adjusted. This setting led to the mean reward of -251 and the best reward of -112 for the pendulum problem.
\end{itemize}
Results of the first three sets of experiments on different environments show the competitiveness of the proposed algorithm compared to the state-of-the-art algorithms and provide a proof of concept. In AO2, the level of abstractness of each option is not determined by any structural limitation. In other words, the system itself is responsible for the depth of each schema. The additional sets of experiments on behavioral analysis were focused on interpretability of the proposed architecture. Each one of these experiments aimed at establishing a correspondence between one of the aspects of human cognition with a characteristic of the proposed algorithm. For instance, $W$ implements a simple attention mechanism in the proposed structure.

\section{Conclusion}

The language of thought hypothesis was proposed as a complementary approach to address some of the shortcomings of the connectionist viewpoint. Both cognitive development and schema theory are promising ideas that can lead to the goals of LOTH. An architecture based on the schema theory is able to take at least one step towards achieving human-like cognition in AI systems. Concepts such as events, objects, and time as well as their relationships with perception can be naturally defined in a schema-based framework, which most of the established models might not be able to fully address. The simulation results for the three case studies demonstrate that such a framework has the potential to solve real-world RL problems in a completely interpretable manner without using neural networks. This can be viewed as a step towards an alternative approach to the mainstream deep RL research, which calls for further investigation. The proposed architecture has achieved better results compared to other algorithms in some environments while it is also able to cover many established theories in cognitive psychology that will improve the learning agent's functionality. Future work will reveal more capabilities of this architecture such as event perception, different views of a single event, more sophisticated perceptual cycle models, and integration of top-down and bottom-up learning. Although, in AO2, the abstraction level is determined by the system itself, and each option may have a different depth, all the decision paths of the AO2 is human-interpretable with respect to the environment state.

\bibliographystyle{IEEEtran}
\bibliography{IEEEabrv,Main}

\begin{thebibliography}{10}
\providecommand{\url}[1]{#1}
\csname url@samestyle\endcsname
\providecommand{\newblock}{\relax}
\providecommand{\bibinfo}[2]{#2}
\providecommand{\BIBentrySTDinterwordspacing}{\spaceskip=0pt\relax}
\providecommand{\BIBentryALTinterwordstretchfactor}{4}
\providecommand{\BIBentryALTinterwordspacing}{\spaceskip=\fontdimen2\font plus
\BIBentryALTinterwordstretchfactor\fontdimen3\font minus
  \fontdimen4\font\relax}
\providecommand{\BIBforeignlanguage}[2]{{%
\expandafter\ifx\csname l@#1\endcsname\relax
\typeout{** WARNING: IEEEtran.bst: No hyphenation pattern has been}%
\typeout{** loaded for the language `#1'. Using the pattern for}%
\typeout{** the default language instead.}%
\else
\language=\csname l@#1\endcsname
\fi
#2}}
\providecommand{\BIBdecl}{\relax}
\BIBdecl

\bibitem{fodor1988connectionism}
J.~A. Fodor, Z.~W. Pylyshyn \emph{et~al.}, ``Connectionism and cognitive
  architecture: A critical analysis,'' \emph{Cognition}, vol.~28, no. 1-2, pp.
  3--71, 1988.

\bibitem{aydede2010language}
M.~Aydede, ``The language of thought hypothesis,'' \emph{Stanford encyclopedia
  of philosophy}, 2010.

\bibitem{piaget2008psychology}
J.~Piaget and B.~Inhelder, \emph{The Psychology of the Child}.\hskip 1em plus
  0.5em minus 0.4em\relax Basic books, 2008.

\bibitem{anderson2000cognitive}
J.~R. Anderson, \emph{Cognitive Psychology and Its Implications}.\hskip 1em
  plus 0.5em minus 0.4em\relax Worth publishers, 2000.

\bibitem{hampson1996understanding}
P.~J. Hampson and P.~E. Morris, \emph{Understanding Cognition}.\hskip 1em plus
  0.5em minus 0.4em\relax Blackwell Oxford, 1996.

\bibitem{skinner2011behaviorism}
B.~F. Skinner, \emph{About Behaviorism}.\hskip 1em plus 0.5em minus 0.4em\relax
  Vintage, 2011.

\bibitem{gregory1970intelligent}
R.~L. Gregory, ``The intelligent eye.'' 1970.

\bibitem{gibson1966senses}
J.~J. Gibson, ``The senses considered as perceptual systems,'' 1966.

\bibitem{sun2004top}
R.~Sun and X.~Zhang, ``Top-down versus bottom-up learning in cognitive skill
  acquisition,'' \emph{Cognitive Systems Research}, vol.~5, no.~1, pp. 63--89,
  2004.

\bibitem{hanson1996development}
C.~Hanson and S.~J. Hanson, ``Development of schemata during event parsing:
  Neisser's perceptual cycle as a recurrent connectionist network,''
  \emph{Journal of Cognitive Neuroscience}, vol.~8, no.~2, pp. 119--134, 1996.

\bibitem{niesser1976cognition}
U.~Niesser, \emph{Cognition and Reality: Principles and Implications of
  Cognitive Psychology}.\hskip 1em plus 0.5em minus 0.4em\relax Freeman, San
  Francisco, 1976.

\bibitem{cognitivecontrol2012}
S.~Haykin, M.~Fatemi, P.~Setoodeh, and Y.~Xue, ``Cognitive control,''
  \emph{Proceedings of the IEEE}, vol. 100, no.~12, pp. 3156--3169, 2012.

\bibitem{fatemi2017observability}
M.~Fatemi, P.~Setoodeh, and S.~Haykin, ``Observability of stochastic complex
  networks under the supervision of cognitive dynamic systems,'' \emph{Journal
  of Complex Networks}, vol.~5, no.~3, pp. 433--460, 2017.

\bibitem{sutton1999between}
R.~S. Sutton, D.~Precup, and S.~Singh, ``Between {MDP}s and semi-{MDP}s: A
  framework for temporal abstraction in reinforcement learning,''
  \emph{Artificial intelligence}, vol. 112, no. 1-2, pp. 181--211, 1999.

\bibitem{precup2001temporal}
D.~Precup, ``Temporal abstraction in reinforcement learning,'' 2001.

\bibitem{mcgovern2001automatic}
A.~McGovern and A.~G. Barto, ``Automatic discovery of subgoals in reinforcement
  learning using diverse density,'' 2001.

\bibitem{menache2002q}
I.~Menache, S.~Mannor, and N.~Shimkin, ``Q-cut—dynamic discovery of sub-goals
  in reinforcement learning,'' in \emph{European Conference on Machine
  Learning}.\hskip 1em plus 0.5em minus 0.4em\relax Springer, 2002, pp.
  295--306.

\bibitem{silver2012compositional}
D.~Silver and K.~Ciosek, ``Compositional planning using optimal option
  models,'' \emph{arXiv preprint arXiv:1206.6473}, 2012.

\bibitem{csimcsek2009skill}
{\"O}.~{\c{S}}im{\c{s}}ek and A.~G. Barto, ``Skill characterization based on
  betweenness,'' in \emph{Advances in neural information processing systems},
  2009, pp. 1497--1504.

\bibitem{stolle2002learning}
M.~Stolle and D.~Precup, ``Learning options in reinforcement learning,'' in
  \emph{International Symposium on abstraction, reformulation, and
  approximation}.\hskip 1em plus 0.5em minus 0.4em\relax Springer, 2002, pp.
  212--223.

\bibitem{daniel2016probabilistic}
C.~Daniel, H.~Van~Hoof, J.~Peters, and G.~Neumann, ``Probabilistic inference
  for determining options in reinforcement learning,'' \emph{Machine Learning},
  vol. 104, no. 2-3, pp. 337--357, 2016.

\bibitem{konidaris2011autonomous}
G.~Konidaris, S.~Kuindersma, R.~Grupen, and A.~Barto, ``Autonomous skill
  acquisition on a mobile manipulator,'' in \emph{Twenty-Fifth AAAI Conference
  on Artificial Intelligence}, 2011.

\bibitem{kulkarni2016hierarchical}
T.~D. Kulkarni, K.~Narasimhan, A.~Saeedi, and J.~Tenenbaum, ``Hierarchical deep
  reinforcement learning: Integrating temporal abstraction and intrinsic
  motivation,'' in \emph{Advances in neural information processing systems},
  2016, pp. 3675--3683.

\bibitem{mankowitz2016adaptive}
D.~J. Mankowitz, T.~A. Mann, and S.~Mannor, ``Adaptive skills adaptive
  partitions ({ASAP}),'' in \emph{Advances in Neural Information Processing
  Systems}, 2016, pp. 1588--1596.

\bibitem{mann2015approximate}
T.~A. Mann, S.~Mannor, and D.~Precup, ``Approximate value iteration with
  temporally extended actions,'' \emph{Journal of Artificial Intelligence
  Research}, vol.~53, pp. 375--438, 2015.

\bibitem{niekum2013semantically}
S.~D. Niekum, ``Semantically grounded learning from unstructured
  demonstrations,'' 2013.

\bibitem{vezhnevets2016strategic}
A.~Vezhnevets, V.~Mnih, S.~Osindero, A.~Graves, O.~Vinyals, J.~Agapiou
  \emph{et~al.}, ``Strategic attentive writer for learning macro-actions,'' in
  \emph{Advances in neural information processing systems}, 2016, pp.
  3486--3494.

\bibitem{bacon2017option}
P.-L. Bacon, J.~Harb, and D.~Precup, ``The option-critic architecture,'' in
  \emph{Thirty-First AAAI Conference on Artificial Intelligence}, 2017.

\bibitem{harb2018waiting}
J.~Harb, P.-L. Bacon, M.~Klissarov, and D.~Precup, ``When waiting is not an
  option: Learning options with a deliberation cost,'' in \emph{Thirty-Second
  AAAI Conference on Artificial Intelligence}, 2018.

\bibitem{klissarov2017learnings}
M.~Klissarov, P.-L. Bacon, J.~Harb, and D.~Precup, ``Learnings options
  end-to-end for continuous action tasks,'' \emph{arXiv preprint
  arXiv:1712.00004}, 2017.

\bibitem{botvinick2008hierarchical}
M.~M. Botvinick, ``Hierarchical models of behavior and prefrontal function,''
  \emph{Trends in cognitive sciences}, vol.~12, no.~5, pp. 201--208, 2008.

\bibitem{nachum2018near}
O.~Nachum, S.~Gu, H.~Lee, and S.~Levine, ``Near-optimal representation learning
  for hierarchical reinforcement learning,'' \emph{arXiv preprint
  arXiv:1810.01257}, 2018.

\bibitem{riemer2018learning}
M.~Riemer, M.~Liu, and G.~Tesauro, ``Learning abstract options,'' in
  \emph{Advances in Neural Information Processing Systems}, 2018, pp.
  10\,424--10\,434.

\bibitem{dayan1993improving}
P.~Dayan, ``Improving generalization for temporal difference learning: The
  successor representation,'' \emph{Neural Computation}, vol.~5, no.~4, pp.
  613--624, 1993.

\bibitem{stachenfeld2014design}
K.~L. Stachenfeld, M.~Botvinick, and S.~J. Gershman, ``Design principles of the
  hippocampal cognitive map,'' in \emph{Advances in neural information
  processing systems}, 2014, pp. 2528--2536.

\bibitem{schulman2017proximal}
J.~Schulman, F.~Wolski, P.~Dhariwal, A.~Radford, and O.~Klimov, ``Proximal
  policy optimization algorithms,'' \emph{arXiv preprint arXiv:1707.06347},
  2017.

\bibitem{zhang2019dac}
S.~Zhang and S.~Whiteson, ``{DAC}: The double actor-critic architecture for
  learning options,'' in \emph{Advances in Neural Information Processing
  Systems}, 2019, pp. 2010--2020.

\end{thebibliography}

\end{document}